\documentclass[final]{cvpr}

\usepackage{times}
\usepackage{epsfig}
\usepackage{graphicx}
\usepackage{amsmath}
\usepackage{amssymb}
\usepackage{multirow}
\usepackage{multicol}
\usepackage{bm}

\newcommand{\y}{\mathbf{y}}
\newcommand{\z}{\mathbf{z}}
\newcommand{\target}{{\bm\tau}}
\DeclareMathOperator{\surrogateloss}{\mathcal{L}^\mathrm{sur}}
\DeclareMathOperator{\targetloss}{\mathcal{L}^\mathrm{sur}_\target}
\DeclareMathOperator{\sceloss}{\mathcal{L}^\mathrm{sce}}
\def \zz {\mathbf{z}}
\def \xx {\mathbf{x}}

\usepackage[pagebackref=true,breaklinks=true,colorlinks,bookmarks=false]{hyperref}

\def\cvprPaperID{****} 
\def\confYear{CVPR 2021}

\begin{document}

\title{Adversarial Attacks on ML Defense Models Competition}

\author{
Yinpeng Dong$^{1,3}$, Qi-An Fu$^{1}$, Xiao Yang$^{1}$, Wenzhao Xiang$^{4}$, Tianyu Pang$^{1}$, Hang Su$^{1}$, Jun Zhu$^{1,3}$, \\Jiayu Tang$^3$, Yuefeng Chen$^{2}$, XiaoFeng Mao$^{2}$, Yuan He$^{2}$, Hui Xue$^{2}$, Chao Li$^{2}$,
\\Ye Liu$^{5}$, Qilong Zhang$^{5}$, Lianli Gao$^{5}$, Yunrui Yu$^{6}$, Xitong Gao$^{7}$,  Zhe Zhao$^{8}$, Daquan Lin$^{8}$,\\Jiadong Lin$^{9}$, Chuanbiao Song$^{9}$, Zihao Wang$^{10}$, Zhennan Wu$^{10}$, Yang Guo$^{11}$, \\ Jiequan Cui$^{12}$, Xiaogang Xu$^{12}$, Pengguang Chen$^{12}$ \vspace{0.3cm}\\

$^{1}$ Tsinghua University \; $^{2}$ Alibaba Group \; $^{3}$ RealAI \; $^{4}$ Shanghai Jiao Tong University\\
$^{5}$ University of Electronic Science and Technology of China \; $^{6}$ University of Macau  \\
$^{7}$ Chinese Academy of Sciences  \; $^{8}$ ShanghaiTech University  \\
$^{9}$ Huazhong University of Science and Technology \; $^{10}$ Indiana University Bloomington \\ $^{11}$ University of Wisconsin–Madison \;
$^{12}$  The Chinese University of Hong Kong \\

\small{\{dyp17, qaf19, yangxiao19, pty17\}@mails.tsinghua.edu.cn}, \small{\{suhangss, dcszj\}@tsinghua.edu.cn} \\
\small{\{yuefeng.chenyf, mxf164419, heyuan.hy, hui.xueh, lizhao.lz\}@alibaba-inc.com}
}

\maketitle

\begin{abstract}
   Due to the vulnerability of deep neural networks (DNNs) to adversarial examples, a large number of defense techniques have been proposed to alleviate this problem in recent years. However, the progress of building more robust models is usually hampered by the incomplete or incorrect robustness evaluation. 
   To accelerate the research on reliable evaluation of adversarial robustness of the current defense models in image classification, the TSAIL group at Tsinghua University and the Alibaba Security group organized this competition along with a CVPR 2021 workshop on adversarial machine learning (\url{https://aisecure-workshop.github.io/amlcvpr2021/}). The purpose of this competition is to motivate novel attack algorithms to evaluate adversarial robustness more effectively and reliably. The participants were encouraged to develop stronger white-box attack algorithms to find the worst-case robustness of different defenses. This competition was conducted on an adversarial robustness evaluation platform --- ARES (\url{ https://github.com/thu-ml/ares}), and is held on the TianChi platform (\url{https://tianchi.aliyun.com/competition/entrance/531847/introduction}) as one of the series of AI Security Challengers Program. After the competition, we summarized the results and established a new adversarial robustness benchmark at \url{https://ml.cs.tsinghua.edu.cn/ares-bench/}, which allows users to upload adversarial attack algorithms and defense models for evaluation.
\end{abstract}


\section{Introduction}

Despite the remarkable success of deep neural networks (DNNs) in various applications~\cite{Goodfellow-et-al2016}, these models are vulnerable to adversarial examples~\cite{carlini2017towards,dong2018boosting,goodfellow2014explaining,Kurakin2016,szegedy2013intriguing}, which are maliciously generated by adding imperceptible perturbations to normal examples but can cause erroneous predictions.
As the vulnerability of DNNs raises concerns in various security-sensitive applications, a large number of adversarial defense methods have been proposed, including randomization~\cite{cohen2019certified,xie2017mitigating}, image denoising~\cite{liao2018defense}, ensemble learning~\cite{pang2019improving,tramer2017ensemble}, adversarial detection~\cite{lee2018simple,pang2018towards}, and adversarial training~\cite{goodfellow2014explaining,madry2017towards}, and certified defenses~\cite{wong2018provable}. 
However, most of the defenses have soon been shown to be ineffective against stronger or adaptive attacks~\cite{Athalye2018Obfuscated,dong2019evading,tramer2020adaptive,uesato2018adversarial,yang2021boosting}, making it challenging to understand the effects of the current defenses and identify the progress of the field. Among these defenses, adversarial training is arguably the most effective approach~\cite{Athalye2018Obfuscated,dong2020benchmarking}, which trains the network on the adversarial examples generated by different attacks instead of the natural examples~\cite{madry2017towards}.

The most widely adopted approach to evaluate adversarial robustness is using adversarial attacks. One of the most popular attacks is the \emph{projected gradient descent} (PGD) method~\cite{madry2017towards}, which iteratively generates an adversarial example by performing gradient updates to maximize a classification loss (e.g., cross-entropy loss) w.r.t. input.
Based on PGD, recent improvements have been made in different aspects, including using different loss functions~\cite{carlini2017towards,croce2020reliable,gowal2020alternative}, adjusting the step size during adversarial attacks~\cite{croce2020reliable}, and introducing new strategies of initialization~\cite{yusuke2020diversity}. 
For example, AutoAttack~\cite{croce2020reliable} establishes a stronger baseline for reliable evaluation of adversarial robustness, which is composed of an ensemble of four attack methods, including APGD\textsubscript{CE}, APGD\textsubscript{DLR}, FAB, and Square Attack. 
The output diversified initialization (ODI)~\cite{yusuke2020diversity} method proposes to improve the diversity of adversarial examples in the output space by adopting a new initialization strategy. 

\begin{table*}[t]
\small
    \centering
\begin{tabular}{l|c|c||l|c|c}
\hline
\multicolumn{3}{c||}{\textbf{Stage I}} 
& \multicolumn{3}{c}{\textbf{Stage II \& Final Stage}} \\
Defense Model & Architecture & Dataset & Defense Model & Architecture & Dataset \\
\hline
Madry et al. (2018) \cite{madry2017towards} & WRN-28-10 & CIFAR-10 & Wang et al. (2020) \cite{Wang2020Improving} & WRN-28-10 & CIFAR-10 \\
Carmon et al. (2019) \cite{carmon2019unlabeled} & WRN-28-10 & CIFAR-10 & Ding et al. (2020) \cite{ding2019mma} & WRN-28-10 & CIFAR-10 \\
Zhang et al. (2019) \cite{zhang2019theoretically} & WRN-34-10 & CIFAR-10 & Wang \& Zhang (2019) \cite{wang2019bilateral} & WRN-28-10 & CIFAR-10 \\
Zhang \& Wang (2019) \cite{zhang2019defense} & WRN-28-10 & CIFAR-10 & Huang et al. (2020) \cite{huang2020self} & WRN-34-10 & CIFAR-10 \\
Hendrycks et al. (2019) \cite{hendrycks2019using} & WRN-28-10 & CIFAR-10 & Chen et al. (2020) \cite{chen2020adversarial} & (3$\times$) ResNet-50 & CIFAR-10 \\
Pang et al. (2020a) \cite{pang2020boosting} & WRN-34-10 & CIFAR-10 & Gowal et al. (2020) \cite{gowal2020uncovering} & WRN-28-10 & CIFAR-10 \\
Rice et al. (2020) \cite{rice2020overfitting} & WRN-34-10 & CIFAR-10 & Alayrac et al. (2019) \cite{alayrac2019labels} & WRN-106-8 & CIFAR-10 \\
Pang et al. (2020b) \cite{pang2019rethinking} & ResNet-v2-47 & CIFAR-10 & Dong et al. (2020) \cite{dong2020adversarial} & WRN-28-10 & CIFAR-10 \\
Wong et al. (2020) \cite{wong2020fast} & ResNet-18 & CIFAR-10 & Pang et al. (2021) \cite{pang2020bag} & DenseNet-121 & CIFAR-10 \\
Shafahi et al. (2019) \cite{shafahi2019adversarial} & WRN-34-10 & CIFAR-10 & Pang et al. (2021) \cite{pang2020bag} & DenseNet-201 & CIFAR-10 \\
Wu et al. (2020) \cite{wu2020adversarial} & WRN-28-10 & CIFAR-10 & Pang et al. (2021) \cite{pang2020bag} & DPN & CIFAR-10 \\
Sehwag et al. (2020) \cite{sehwag2020hydra} & WRN-28-10 & CIFAR-10 & Pang et al. (2021) \cite{pang2020bag} & WRN-34-10 & CIFAR-10 \\
Pang et al. (2021) \cite{pang2020bag} & WRN-34-10 & CIFAR-10 & Pang et al. (2021) \cite{pang2020bag} & WRN-34-10 & CIFAR-10  \\
Wong et al. (2020) \cite{wong2020fast} & ResNet-50 & ImageNet & Pang et al. (2020a)~\cite{pang2020boosting} & ResNet-50 & ImageNet \\
Shafahi et al. (2019) \cite{shafahi2019adversarial} & ResNet-50 & ImageNet & Xie et al. (2019) \cite{xie2018feature} & ResNet-152 & ImageNet \\
\hline
\end{tabular}
\vspace{1em}
    \caption{The defense models used in the competition.}
    \label{tab:models}
\end{table*}

To further accelerate the research on reliable evaluation of adversarial robustness of different defense models, we organized this competition in which the participants were required to develop effective white-box attack algorithms. We focus on image classification models since most adversarial defenses are developed on typical image benchmarks, e.g., CIFAR-10~\cite{krizhevsky2009learning} and ImageNet~\cite{deng2009imagenet}. 
We only consider adversarial training models as the evaluated defenses since 1) these models are the state-of-the-art defenses; 2) these models do not have randomness in their predictions; and 3) the model gradient can be calculated without causing nonexistence of gradients.
The whole competition consisted of three stages. In the first stage, we evaluated the submissions of white-box attacks on 15 defense models, including 13 models on CIFAR-10 and 2 models on ImageNet. We used the first 1,000 images of the CIFAR-10 test set and randomly chose 1,000 images of the ILSVRC 2012 validation set for evaluation. The top-100 teams of the first stage can enter the second one. In the second stage, we evaluated submissions on another set of 15 secret defense models with the same images, to ensure that the submissions cannot adapt to the details of the defense models. The top-20 teams can enter the final stage, in which we evaluated the final submissions using the whole CIFAR-10 test set containing 10,000 images and randomly chosen 10,000 images from the ILSVRC 2012 validation set. We evaluated the submissions of white-box attacks by the misclassification rate of the defense models (higher is better).
Below we present more details about the competition and the top-6 solutions.

\section{Competition Details}

This competition was conducted on an adversarial robustness evaluation platform --- ARES (\url{https://github.com/thu-ml/ares})~\cite{dong2020benchmarking}. The participants needed to implement white-box attack algorithms following the \emph{Attacker} class in ARES. The whole competition consisted of three stages with different models and datasets.

\subsection{Models and Datasets}

\textbf{Stage I.} In the first stage, we evaluated the submissions of white-box attacks on 15 defense models, which include 13 models trained on CIFAR-10 and 2 models trained on ImageNet. The models are shown in Table~\ref{tab:models}.
We used the first 1,000 images from the CIFAR-10 test set and randomly chose 1,000 images from the ILSVRC 2012 validation set for evaluating the submissions on the public leaderboard.

\textbf{Stage II.} The top-100 teams in Stage I can enter Stage II. In this stage, we evaluated submissions on another set of 15 secret models. The models are shown in Table~\ref{tab:models}. The datasets are the same as Stage I.

\textbf{Final Stage.} We evaluated the final scores of the top-20 teams in Stage II. Each participant can select two submissions for the final evaluation. The models are the same as those used in Stage II. We used all the 10,000 images from the CIFAR-10 test set and randomly chosen 10,000 images from the ILSVRC 2012 validation set for final evaluations.

\subsection{Evaluation}
The participants were required to submit the source code of their developed white-box attacks. The submissions were evaluated under the untargeted $\ell_\infty$-norm threat models. For CIFAR-10, the perturbation budget is $\epsilon=8/255$. For ImageNet, the perturbation budget is $\epsilon=4/255$.
The submitted attacks have access to the white-box models, but are limited to using their logit outputs. The participants can design losses and attack algorithms based on the logits for gradient calculations. We encouraged the participants to develop ``general'' attack algorithms, which means that they are not specified to each model.
We evaluated the submitted attacks by the misclassification rate of the defenses models (higher is better), which is computed by the following formula
\begin{equation}
    Score(A)=\frac{1}{|\mathcal{M}|}\sum_{M_i\in\mathcal{M}}\frac{1}{|\mathcal{D}|}\sum_{(x_j,y_j)\in\mathcal{D}}\mathbf{1}(M_i(A(x_j))\neq y_j),
\end{equation}
where $A$ is an attack method that returns the adversarial example given a natural one, $M_i$ is a defense model, and $(x_j, y_j)$ is an image-label pair.
The participants needed to ensure that $(A(x_j))$ returns an adversarial example whose distance from $x_j$ is smaller than $\epsilon$, otherwise we would clip the adversarial example within the range.

\subsection{Additional Restrictions}

To save computational cost and make a fair comparison between different attack methods, we restricted the runtime of the attack submissions. The average number of backward gradient calculations per image should be less than 100; the average number of forward model predictions should be less than 200. The total runtime of a submission for all models should be less than 3 hours on a Tesla V100 GPU (as a comparison, the baseline PGD-100 attack needs less than 2 hours for all models).
When the attack runs backward gradient calculation, running forward model prediction is necessary, such that each backward gradient calculation would consume one backward gradient calculation quota and one forward model prediction quota, even if the attack does not explicit run the forward model predictions. 
\begin{table}[t]
\small
    \centering
\begin{tabular}{c|c|c|c}
\hline

Team
& \textbf{Stage I} 
& \textbf{Stage II} & \textbf{Final Stage} \\
\hline

green hand & \bf52.53 & \bf50.93 & \bf51.104 \\ 
\hline
UM-SIAT & 52.39 & 50.65 & 50.961 \\ 
\hline
S3L & 52.13 & 50.56 & 50.955 \\
\hline
BorderLine & 52.40 & 50.69 & 50.895 \\
\hline
Kanra & 52.18 & 50.65 & 50.888 \\
\hline
BalaBala2020 & 51.89 & 50.55 & 50.887 \\
\hline
\end{tabular}
\vspace{1em}
    \caption{Competition results of the top-6 teams.}
    \label{tab:analysis_other}
\end{table}

\section{Competition Results}

There are more than 1,600 teams participating the competition. The total number of submissions is over 2,500. We received about 100 high-quality attack algorithms during the competition. The scores of the top-6 submissions are shown in Table~\ref{tab:analysis_other}. The final scores of the top-6 submissions are very close. They all lie within $[50.887,51.104]$. After the competition, we summarized the results and established a new adversarial robustness benchmark at \url{https://ml.cs.tsinghua.edu.cn/ares-bench/}. This benchmark includes several typical attacks (e.g., PGD-100, MIM-100, CW-100) as well as the top-5 attacks in the competition to evaluate adversarial robustness of different defense models. This benchmark also allows users to upload adversarial attacks and defenses for evaluation. We will further maintain the benchmark results after the competition for evaluating and comparing future attack or defense methods. 

\section{Top Scoring Submissions}

\subsection{1st place: green hand}
\textbf{Team members}: Ye Liu, Qilong Zhang, Lianli Gao

\subsubsection {Method}
   
   

Since the numbers of backward gradient calculations and the forward model predictions are limited, we need to make effective use of it. We adopt ODI-PGD~\cite{yusuke2020diversity} as the baseline method and make several improvements upon it.

\textbf{Step decay}: The fixed step size is sub-optimal, so we gradually reduce the step size from the maximum to one hundred-th of the maximum step size during the attack. Specifically, we use the SGDR~\cite{loshchilov2016sgdr} method in the step adjustment strategy. In each restart, we utilize a large step size at the beginning and then quickly reduce the step size so that we can converge to local maximum of loss. 
The overall step size can be formalized as: 
\begin{equation}
    \begin{split} 
    \eta_i =&  \frac{1}{2}(\eta_{max}-\eta_{min})\left( 1+\cos\left( \frac{i\ \mathrm{mod} \ T}{T} \pi \right) \right)+\eta_{min},
    \label{eq_4} 
    \end{split}
\end{equation}
where $\eta _i$ indicates the step size of the $i$-th iteration, $\eta _{min}$ represents the smallest step size, $\eta _{max}$ represents the maximum step size, $i$ denotes the current iteration number, and $T$ represents the total number of iterations.

\textbf{Delete difficult examples}: The attack difficulty of different examples is different, and we cannot successfully attack all examples. Therefore, we do not attack the difficult examples. Specifically, we judge the difficulty of the examples based on the loss. If the loss is relatively small, we think the example is more difficult to attack, if the loss is relatively large, we think the example is easier to attack successfully.

\textbf{Iteration increase}: In order to efficiently use the number of iterations, we use multiple restarts. At the beginning, the number of iterations for restart is smaller. As the number of restarts increases, the number of iterations included in  restart also increases.

\textbf{Bias output diversity initialization}: The output diversity initialization method is moving in a random direction. We propose a biased output diversity initialization method, which can move in a better direction to improve the efficiency of the attack.

\subsubsection{Submission Details and Results}

\textbf{Submission details}: The number of restarts is 17, the number of iterations is [10-60], maximum step size $\eta _{max}$ is 8/255, smallest step size $\eta _{min}$ is $0.001 \times \eta _{max}$.

\noindent \textbf{Results}: The preliminary score is 52.53, the semifinal score is 50.93, the final score is 51.104. All scores are ranked first.

\subsection{2nd place: team UM-SIAT}
\textbf{Team members}: Yunrui Yu, Xitong Gao

\subsubsection{Method}

We used the new surrogate loss function
introduced in our earlier work~\cite{yu2021lafeat},
which can improve attack success rate and convergence rate
with this function
when compared to the original loss function
used to train the model.
For this competition,
we divided the model into
robust models and non-robust ones,
and designed different strategies for each type.
In order to maximize the score,
we additionally explored
the use of step-size schedules,
ensemble strategies,
and adapts the number of iterations
for different image samples.
The next section will explain
our method in detail.

\subsubsection{Submission Details and Results}

We found that
for many of the defense models,
the output from the standard
softmax cross-entropy loss function
is often easily saturated,
\ie, it is prone to underflow in floating-point arithmetic,
or produce negligible back-propagation signals.
In some cases,
attacks based on back-propagation on this function will fail.
To overcome this,
we used the LAFEAT loss function introduced
in our earlier work~\cite{yu2021lafeat},
which constrains the range of the difference
between the maximum and the second largest value
in the logits output of the model:
\begin{equation}
    \surrogateloss{(\z, \y)}
    \triangleq \sceloss\!\left(
        {\z} \big/ {\left(
            \y^\top \z -
            \max\left(
                \left( 1 - \y \right) \cdot \z
            \right)
        \right)}, \y
    \right),
    \label{eq:surrogate}
\end{equation}
where \( \z \) is the pre-logits network output,
\( \y \) is the one-hot ground-truth label,
and \( \cdot \) denotes the element-wise product.
Using this loss function,
gradient-based attacks can in general
converge significantly faster,
and improve the success rate of attacks.

The second tactic we adopted
in our method is targeted variants
of the above loss function.
Instead of maximizing
the loss of the ground-truth label,
the targeted variant minimizes the loss of the target label:
\begin{equation}
    \targetloss(\z, \y) \triangleq -\sceloss\!\left(
        \frac\z{t} {\left(
            \y^\top \z -
            \max\left(
                \left( 1 - \y \right) \cdot \z
            \right)
        \right)}^{-1}, \target
    \right).
    \label{eq:surrogate:target}
\end{equation}
We found that the targeted variants
could further reduce the accuracy
of the non-robust models.
Based on this finding,
we designed a simple tactic
to automatically distinguish
if the model is robust against our baseline strategy,
and adjust our strategy accordingly.
We designed a 10-iteration attack
which uses the untargeted surrogate loss,
followed by the top-3 targets
of the DLR loss~\cite{croce2020reliable}.
We classify models as non-robust,
if the robustness of the model
is dropped sharply during the targeted phase.
We divide the remaining number
of gradient-iteration budget by 9
to attack the non-robust model
by the top-9 targets of the new loss function.
We will use all remaining iterations
to attack robust models
using the untargeted loss function.
As the different attack settings
are carried out in sequence,
we allow new attack settings to
continue working on the previous perturbation
generated by the previous setting.

We explored different choices
of step size schedules,
and used \( \cos(\frac{4i}I) \)
as the final version,
where \( i \) is the current iteration count
and \( I \) is the number of iterations
of the current attack.
After each attack iteration,
if the sample has been attacked successfully,
we will stop further attacks on the sample
and allocate the remaining iterations
to outstanding samples
that have not been attacked successfully.

To ensure that the attack strategy
can generalize to new models,
we search for the optimal configuration
with results that are relatively stable
under local hyperparameter changes.


\paragraph{Result:} The final score is 50.961.

\subsection{3rd place: team S3L}
\textbf{Team members}: Zhe Zhao and Daquan Lin

\subsubsection{Method}
\label{thirdmethod}
The critical point of this competition is the limitation on queries,
which requires the attacker to construct more efficient attack methods.
We conducted a simple test of each adversarially trained model using PGD attack, 
and found that these models often have similar characteristics when facing an adversarial attack: 
about 50\% of the benign samples can be easily attacked successfully,
about 40\% of the samples are almost impossible to find adversarial examples,
the remaining 10\% are the key to compete with other teams, 
and these images have adversarial examples, 
but require a lot of iterations to find them.

We designed our method for this competition from two perspectives.
\begin{enumerate}
    \item For images that are easy to attack: finding adversarial examples using as few iterations as possible.
    \item For images that cannot be attacked, finding and excluding them using as few iterations as possible.
\end{enumerate}

The key point of our method is to estimate \textit{local robustness} effectively and efficiently. 

\newtheorem{definition}{Definition}[section]
\begin{definition}[Local Robustness]
Given a sample input $x$, a DNN $\mathcal{D}$ and a perturbation threshold $\epsilon$, $\mathcal{D}$ is $\epsilon$ -local robust iff for any sample input $x^{\prime}$ such that $\left\|x-x^{\prime}\right\|_{p} \leq \delta$, we have $\mathcal{D}(x)=\mathcal{D}\left(x^{\prime}\right)$, where $\|\cdot\|_{p}$ is the
$p$ -norm to measure the distance between two images.
\end{definition}

There are some techniques based on software analysis and verification
(e.g., interval analysis,
abstract interpretation, etc.~\cite{huang2020survey, zhang2021bdd4bnn}) 
that can derive the robust accuracy of a neural network soundly.
The robust accuracy refers to the accuracy that 
a neural network can achieve under the local robustness constraint.
In this competition the models are white-box models, 
but the organizers restrict the attackers to use only gradient data.
Thus we cannot use the verification techniques mentioned above. 
Therefore, we propose a novel attack method, 
named outside-inside attack~(OIA), 
to estimate whether an input has the potential to be successfully attacked.\footnote{The paper based on OIA is in preparation and will be available soon.}
The inspiration of OIA comes from~\cite{zhao2021attack},
which uses attack costs to estimate local robustness.

OIA first attacks the benign samples using a perturbation larger than $\epsilon$
(e.g., $2 \times \epsilon$),
and subsequently projects the perturbation to $\epsilon$ under $L_\infty$ constraint.
In the first step, 
if the image is successfully attacked with a larger perturbation, 
we consider it to have an adversarial example in the $\epsilon$ interval.
While for a failed attack, 
we think the input has no adversarial example in the $\epsilon$ interval, i.e., satisfying local robustness.
For these images, 
we remove them from the subsequent iterations, 
thus saving the iterations for those images 
that are likely to be successfully attacked.

\subsubsection{Submission Details and Results}
For OIA, we used a 5-step BIM attack with $\alpha = \epsilon/2$.
Subsequently, we used ODI-PGD attack~\cite{yusuke2020diversity}.
Referring to the hyperparameters provided by ~\cite{yusuke2020diversity}, 
we performed a 2-step ODI attack with $\alpha = \epsilon$.
Then a 20-step PGD search was executed.
It is worth mentioning that 
we use the method recommended in AutoAttack~\cite{croce2020reliable}
to select the optimal perturbation obtained
during the previous attack as the starting point for the next round.

Using OIA before ODI-PGD attack, 
we can optimize the two perspectives mentioned in 
Sec.~\ref{thirdmethod} at the same time.
First, OIA can quickly attack those vulnerable samples 
whose perturbations are also easily effective after clipping.
In addition, OIA uses only a few iterations to
filter out a large number of images that 
cannot be successfully attacked under $L_\infty$ constraint.
However, 
OIA only provides an approximate estimate of local robustness
and cannot give a theoretical guarantee, 
but from our statistical results, 
the false negative rate of this method is only about 0.1\textperthousand.

During the implementation of PGD attack, 
we hard-coded the number of iterations and step size decay. 
In the initial few restarts, 
we use a smaller number of iterations to search quickly,
and in the subsequent restarts, 
we increase the number of iterations. 
During the iterations, 
we keep reducing the step size to avoid oscillations.
Inspired by DFA~\cite{duan2019things}, 
we constrain the minimum step-size in the reducing process 
to avoid accuracy problems.
In the end, we got 50.955 points and took third place.

\subsection{4th place: team BorderLine}
\textbf{Team members}: Jiadong Lin and Chuanbiao Song

\subsubsection{Method}
In this section, we introduce the random real target (RRT) attack method, which won the 4th place in the competition. The idea is to develop a novel sampling strategy for the initial perturbed points to improve the efficiency of adversarial attack.

Random sampling is crucial for the optimization-based white-box attacks, which helps to find a diverse set of initial perturbed points for the attack success. Many sampling strategies have been used to improve the attack efficiency, such as uniform sampling, Gaussian sampling and output diversified sampling (ODS)~\cite{yusuke2020diversity}. Currently, ODS is widely used to randomly restart for the white-box attacks and leads to achieve the best performance among the sampling strategies. The basic idea of ODS is randomly specify a direction in the output space and then perform gradient-based optimization to generate a perturbation in the input space. However in practice, the randomly direction in the output space may not be a great representation of the effective and true direction, which limits the attack strength.

In order to overcome this drawback, we propose random real target (RRT) sampling, a novel sampling strategy that attempts to enhance the sampling diversity in the space of the target model’s outputs while maintain its authenticity. RTT and ODS differ in two ways. First, RRT uses the logits of the real target image as the optimization direction, while ODS uses the random noises sampled from the uniform distribution. Second, RRT uses cosine similarity as the  distance for the  optimization, while ODS directly uses the  sampled direction as the  distance.

Given an original image $x_{ori}$, a random real target image $x_{tar}$, we define the perturbation vector of RRT as follows:
\begin{equation*}
    v_{RRT}(x_{ori}, x_{tar}) = \nabla_{x_{ori}}(\frac{f(x_{ori})\cdot f(x_{tar})}{\| f(x_{ori}) \|_2 * \| f(x_{tar}) \|_2}),
\end{equation*}
where $f$ is a classifier that maps the input image $x\in[0,1]^{D}$ to the logits $z\in\mathbb{R}^C$, $D$ is input dimension and $C$ is the number of classes.

In the competition, we utilize RRT for initialization to generate diversified initial perturbed points and maintain the representative and authenticity of the logits of the initial perturbed points. Following ODS, we perform RRT initialization by the gradient-based optimization. Given an original image $x_{0} = x_{ori}$ and a random real target image $x_{tar}$, we try to find a restart point $x$ via the following iterative update:
\begin{equation*}
\begin{gathered}
    x_{t+1} =\Pi_{\mathcal{B}_{\epsilon}^{\infty}(x_{ori})}\left(x_{t}+\alpha \operatorname{sign}(v_{RRT}(x_{t}, x_{tar}))\right),
\end{gathered}
\end{equation*}
where $\mathcal{B}_{\epsilon}^{\infty}(x) \ :=\left\{x' :\|x'-x\|_{\infty} \leq \epsilon\right\}$ is the set of allowed perturbations, and $\Pi_{\mathcal{B}_{\epsilon}^{\infty}(x)}$ indicates the projection of the set $\mathcal{B}_{\epsilon}^{\infty}(x)$. 
Similar to ODS, we randomly sample a new target image $x_{tar}$ for each random restart for the attack. After obtaining the starting points, we perform the projected gradient descent (PGD) attack to generate adversarial examples.  

\subsubsection{Submission Details and Results}
In this section, we summarize the submission details including hyper-parameters setting and some tricks, and the submission results.

\textbf{Iterative number}. Iterative number is a crucial parameter for adversarial attacks. In this competition, an average of 100 iterations were allowed per sample. We set the iterative number of RRT initialization as 2 and the iterative number of PGD attack as 18.

\textbf{Learning rate}. Following the MD attack method~\cite{jiang2020imbalanced}, we use a large step size ($2 \times \epsilon$) in the first stage (steps 0-4) for a better exploration, and a small step size ($0.25 \cdot \epsilon$) in the second stage (steps 5-17) to ensure a stable optimization.

\textbf{Momentum update}. Following the MIM attack ~\cite{dong2018boosting}, we also integrated the momentum idea into PGD to accelerate the attack. Since in the first stage, the update direction is highly inaccurate, we only perform momentum update in the second stage (step 5-17).

\textbf{Multi-Targeted attack}. Following the MT attack~\cite{gowal2020alternative}, at each restart, we pick a new target class for targeted attack. We first sort the target classes based on the logit outputs, from the highest to the bottom. Then, in turn, the target class is selected to attack until the allowed number of iterations are run out.

\textbf{Submission Results}. As shown in Table~\ref{tab:major}, we report the  major milestone for the changes in attack method. We can see that, at the Stage I, the ODI, MT and MIM strategies can boost the attack success of the PGD attack, and our RRT provides better performance than the ODI, which we can attribute to the diverse and real initial perturbed points.

\begin{table}[ht]
\centering
\begin{tabular}{lllll}
\hline
Scenario & Stage I & Stage II & Final \\
\hline
RRT+PGD+MT+MIM & 52.40 & 50.69 & 50.895 \\
ODI+PGD+MT+MIM & 52.23 & - & - \\
ODI+PGD+MT & 52.09 & - & - \\
ODI+PGD & 51.76 & - & - \\
PGD & 46.18 & - & - \\
\hline
\end{tabular}
\vspace{0.5em}
\caption{Major Millstone for the changes of method}
\label{tab:major}
\end{table}

\subsection{5th place: team Kanra}
\textbf{Team members}: Zihao Wang, Zhennan Wu, Yang Guo

\subsubsection {Method}

Since the average number of backward gradient calculations per image and forward model predictions are restricted, to improve the efficiency of the attack, we propose Fast Restart Projected Gradient Descent (FR-PGD), a two-phase strategy that can efficiently utilize the limited resources.

\begin{figure}[h]
\begin{center}
   \includegraphics[width=0.9\linewidth]{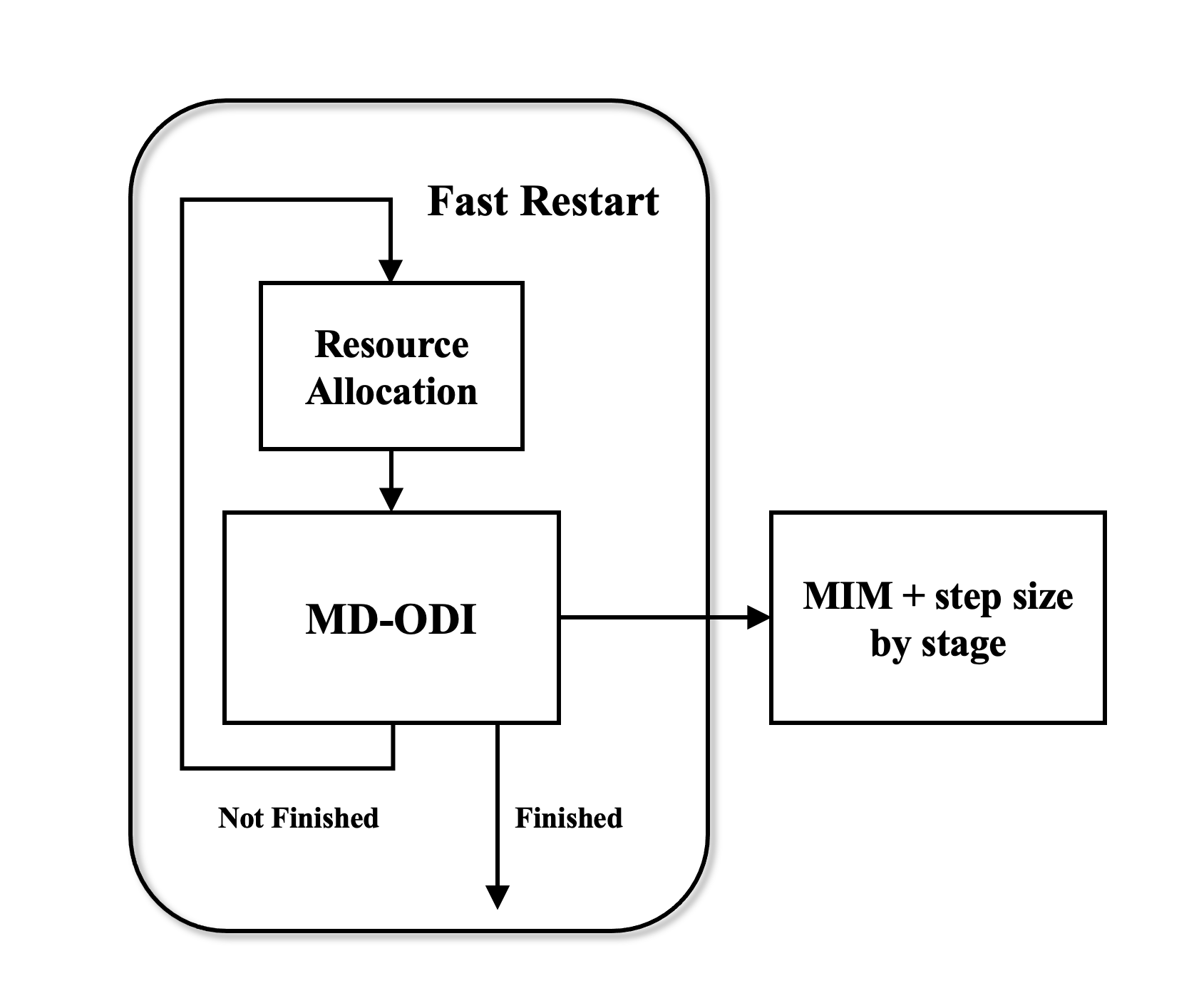}
\end{center}
   \caption{Workflow of Fast Restart Projected Gradient Descent.}
\label{fig:Fast_Restart}
\end{figure}

Figure \ref{fig:Fast_Restart} demonstrates a high-level workflow of our attack. The attack can be divided into two phases: Fast Restart phase and multi-step convergence phase. Compared with previous restart methods, each restart of FR-PGD takes only a small number of backward gradient calculations. Therefore, the number of restarts can be guaranteed even in the case of resource constraints. Then, several adversarial samples from the Fast Restart phase (the number of the samples depends on how many samples have not been successfully attacked in the restart phase) with the highest Loss value are sent to the multi-step convergence phase. Finally, the convergence results are returned.

It is worth noting that although we choose the highest Loss value each time, there is only a positive correlation between the Loss value and the final success probability of the attack to a certain extent, but not absolutely. The more the number of backward gradient calculations in the restart phase, the stronger the correlation. However, this will lead to a reduction in the number of restarts or the steps of convergence, since the resources are limited. Therefore, it needs to be weighed according to resources.

\subsubsection{Submission Details and Results}

Specifically, a Output Diversified Initialization (ODI) \cite{yusuke2020diversity} version of Margin Decomposition (MD) attack \cite{jiang2020imbalanced} is used in the Fast Restart phase. The loss function in each restart is defined as follows:
\begin{align}
\label{eq:ld-pgd}
    \xx_{k+1} &= \Pi_\epsilon (\xx_k + \alpha\cdot\text{sign}(\nabla_{\xx} \ell^r_k(\xx_k, y))),\\
    \ell^r_k(\xx_k,y) &= 
    \begin{cases}
        \zz_{max} & \text{if }k < \frac{K}{2} \text{ and }r \bmod 2 = 0 \\
        -\zz_{y} & \text{if }k < \frac{K}{2} \text{ and } r \bmod 2 = 1 \\
        \zz_{max} -\zz_{y} & \text{if } k \geq \frac{K}{2},
    \end{cases}\nonumber
\end{align}
where, $k \in \{1,\dots,K\}$ is the number of backward gradient calculations in the restart phase, $r \in \{1,\dots,n\}$ is the $r$-th restart, $\bmod$ is the modulo operation for alternating optimization, and $\ell^r_k$ defines the loss function used at the $k$-th step and $r$-th restart. The loss function switches from the individual terms back to the full margin loss at step $\frac{K}{2}$. In each restart, we perform 2 steps of ODI and 4 steps of gradient ascents. After each restart, the computing resources are reallocated. Specifically, the resources of the samples that have been successfully attacked are allocated to the samples that have not been done, so that the unfinished samples can get more number of restarts.

In the multi-step convergence phase, Momentum Iterative Method (MIM) \cite{dong2018boosting} with the C\&W Loss \cite{carlini2017towards} is performed. We adopt scheduled step size instead of fixed one. Because we found that starting from large step size brings better results. We set the initial step size $\eta_0$ as $\eta_0=\epsilon$.
We update the step size to $\epsilon/3, \epsilon/8$ at $k=0.25N, 0.5N$, respectively.

The ratio of resources of the Fast Restart phase and the multi-step convergence phase is set to 4:1 empirically. Our method got a score of 50.888 in the Final Stage.

\subsection{6th place: team BalaBala2020}
\textbf{Team members}: Jiequan Cui, Xiaogang Xu, Pengguang Chen

\subsubsection {Method}
The team proposed a novel attack strategy called \textit{Difficulty-Hierarchical Attack (DH Attack)} for attacking target models with a constrained number of attack iterations.
In this strategy, more attack iterations will be conducted for the robust samples that are harder to perturb, improving the integral attack success rate.
Moreover, a global perturbation \cite{moosavi2017universal} is utilized, attacking a part of samples at the beginning of each attack iteration and allowing more attack iterations for robust samples.
Furthermore, several effective tricks are explored and adopted in DH Attack.

\noindent {\bf Hierarchical Attack.}
To optimally utilize the constrained attack iterations, the team distributes the number of attack iteration to each sample according to their robustness towards adversarial perturbations. The attack is conducted hierarchically with multiple restart times, and once one sample is successfully attacked, the attack process for this sample will be canceled, preserving more attack iterations for robust samples.

\noindent {\bf Global Perturbation.}
Moreover, the team utilizes the transferability of adversarial attacks and creates a global perturbation by accumulating the perturbations of different samples. The creation of global perturbation does not involve the backward of the network, and it can be utilized at the beginning of each restart to attack a part of the samples.

\noindent {\bf Other Tricks.}

\noindent {\bf I.} Loss function ensemble. The team observed that different robust models are sensitive to different loss functions for generating adversarial examples. After each restart with ODI initialization \cite{yusuke2020diversity}, DH Attack would uniformly choose loss function from \{MarginLoss \cite{carlini2017towards}, DLRLoss \cite{croce2020reliable}\}.

\noindent {\bf II.} Decaying step-size. Adopting cosine schedule for step-size is better than a fixed value.

\noindent {\bf III.} ODI initialization with restricted search directions. In the ODI algorithm, the generated initialization with one random vector uniformly exists in the whole search space. It keeps the diversity of initialization at the cost of attack efficiency. In experiments, the team observed that, for most successfully attacked images, the corresponding adversarial examples are misclassified into the first or the second most possible classes finally. Based on this phenomenon, after each ODI initialization, DH Attack would apply 5 attack iterations with a multi-targeted attack loss, {\i.e., } summing the targeted loss for the first and the second most possible classes that could be misclassified into. 

\subsubsection{Submission Details and Results}
For ODI initialization, the team applied 5 iterations to update. The initial step-size is 4/255 and is decayed with a cosine schedule for each restart. In the beginning, the team utilized a 10 iterations attack to filter the easiest samples that can be attacked. We use 20 iterations for all the remaining restart if the backward quota is enough. The team achieved a score of 50.887 for the final testing and rank sixth.

\section {Conclusion}

We successfully organized the \emph{Adversarial Attacks on ML Defense Models} competition, which attracted thousands of teams to participate. We obtained many effective attack algorithms to evaluate adversarial robustness reliably and correctly, which helps to accelerate the research on robustness evaluation. The technical details of the top scoring submissions are presented in this paper. We also established an adversarial robustness benchmark to summarize the results, which could be further used for newly developed attack and defense methods.

{\small
\bibliographystyle{ieee_fullname}
\bibliography{main}
}

\end{document}